\documentclass[letterpaper,12pt]{article}
\usepackage{graphicx}
\usepackage{amssymb,amsmath}
\usepackage[utf8]{inputenc}
\usepackage[T1]{fontenc}
\usepackage{subfigure}
\usepackage{amsthm}
\usepackage{hyperref}
\usepackage{color}
\usepackage{multirow,txfonts,verbatim}
\usepackage{mathrsfs}
\usepackage{booktabs}
\usepackage{bm}
\usepackage{array}
\usepackage{tikz}
\usetikzlibrary{decorations.pathreplacing,calligraphy}
\usepackage{algorithm}
\usepackage{algpseudocode}

\usepackage{float}

\newtheoremstyle{plainNoItalics}{}{}{\normalfont}{}{\bfseries}{.}{ }{}
\theoremstyle{plain}

\theoremstyle{plainNoItalics}

\theoremstyle{definition}
\theoremstyle{definition}
\theoremstyle{definition}
\theoremstyle{definition}
\theoremstyle{definition}
\theoremstyle{definition}
\numberwithin{equation}{section}
\numberwithin{theorem}{section}
\numberwithin{lemma}{section}
\numberwithin{figure}{section}
\numberwithin{example}{section}
\numberwithin{rem}{section}

\usepackage{lineno}
\usepackage{xcolor}
\usepackage{natbib}

\graphicspath{{figures/}}

\title{Machine Unlearning for Robust DNNs: Attribution-Guided Partitioning and Neuron Pruning in Noisy Environments}
\author{
Deliang Jin
\thanks{Corresponding author. School of Computer Science and Artificial Intelligence,Zhengzhou University  {\tt dljin@ha.edu.cn}}
\and
Gang Chen
\thanks{ Zhengzhou University  {\tt chengang@zzu.edu.cn}. }
\and
Shuo Feng
\thanks{ School of Computer Science and Artificial Intelligence,Zhengzhou University {\tt sfeng@ha.edu.cn}}
\and
Yufeng Ling
\thanks{ School of Computer Science and Artificial Intelligence,Zhengzhou University {\tt Iceydying@163.com}}
\and
Haoran Zhu
\thanks{ School of Computer Science and Artificial Intelligence,Zhengzhou University {\tt zhrscd@163.com}}
}

\begin{document}

\maketitle
\begin{abstract}
Deep neural networks (DNNs) have achieved remarkable success across diverse domains, but their performance can be severely degraded by noisy or corrupted training data. Conventional noise mitigation methods often rely on explicit assumptions about noise distributions or require extensive retraining, which can be impractical for large-scale models. Inspired by the principles of machine unlearning, we propose a novel framework that integrates attribution-guided data partitioning, discriminative neuron pruning, and targeted fine-tuning to mitigate the impact of noisy samples. Our approach employs gradient-based attribution to probabilistically distinguish high-quality examples from potentially corrupted ones without imposing restrictive assumptions on the noise. It then applies regression-based sensitivity analysis to identify and prune neurons that are most vulnerable to noise. Finally, the resulting network is fine-tuned on the high-quality data subset to efficiently recover and enhance its generalization performance. This integrated unlearning-inspired framework provides several advantages over conventional noise-robust learning approaches. Notably, it combines data-level unlearning with model-level adaptation, thereby avoiding the need for full model retraining or explicit noise modeling. We evaluate our method on representative tasks (e.g., CIFAR-10 image classification and speech recognition) under various noise levels and observe substantial gains in both accuracy and efficiency. For example, our framework achieves approximately a 10\% absolute accuracy improvement over standard retraining on CIFAR-10 with injected label noise, while reducing retraining time by up to 47\% in some settings. These results demonstrate the effectiveness and scalability of the proposed approach for achieving robust generalization in noisy environments.
\end{abstract}
\section{Introduction}

Deep neural networks (DNNs) have achieved unprecedented success across diverse domains, fundamentally transforming fields such as computer vision~\cite{krizhevsky2012imagenet, he2016deep, dosovitskiy2020image}, natural language processing~\cite{devlin2019bert, vaswani2017attention, brown2020language}, speech recognition~\cite{graves2013speech, hinton2012deep, hannun2014deep}, medical diagnosis~\cite{esteva2017dermatologist, litjens2017survey}, and reinforcement learning~\cite{mnih2015human, silver2016mastering, schulman2017proximal}. These remarkable achievements stem largely from groundbreaking architectural innovations including convolutional neural networks (CNNs) \cite{krizhevsky2012imagenet}, residual networks \cite{he2016deep}, transformer architectures \cite{vaswani2017attention}, and generative adversarial networks (GANs) \cite{goodfellow2014generative}. Each of these developments has contributed substantially to enhanced model performance, training stability, and the ability to capture complex data representations. However, despite these architectural advances, the effectiveness and reliability of DNNs remain fundamentally constrained by the quality and integrity of their training data.

Real-world datasets are inherently susceptible to various forms of noise that compromise both input features and target labels, presenting significant challenges for robust model development. Feature noise manifests through multiple sources including sensor measurement inaccuracies, data acquisition errors, missing or corrupted values, and deliberate adversarial perturbations\cite{biggio2013evasion}  designed to mislead model predictions \cite{frenay2013classification}. Concurrently, label noise  emerges from human annotation inconsistencies, subjective interpretation differences among annotators\cite{northcutt2021pervasive}, ambiguous category boundaries, automated labeling system errors, and temporal changes in labeling criteria \cite{song2019selfie}. The pervasiveness of these noise sources is exemplified in widely-adopted benchmark datasets: comprehensive analysis reveals that approximately 4\% of ImageNet labels contain errors or ambiguities \cite{rolnick2017deep}, while clinical datasets such as MIMIC-III\cite{johnson2016mimic} exhibit substantial missing physiological measurements and inconsistent diagnostic annotations that further complicate reliable model training.

The fundamental challenge lies in the demonstrated capacity of DNNs to memorize arbitrary noise patterns, a phenomenon that severely undermines their generalization capabilities \cite{arpit2017closer}. High-capacity networks can achieve near-zero training error even when trained on datasets with completely randomized labels, indicating their propensity to overfit to spurious correlations rather than learning meaningful underlying patterns \cite{zhang2016understanding,belkin2019reconciling}. This memorization behavior becomes particularly problematic in practical applications where model reliability is critical, as networks may confidently make incorrect predictions based on learned noise patterns rather than genuine signal\cite{amodei2016concrete}.

Existing approaches to address noise-related challenges in deep learning can be broadly categorized into several paradigms, each with distinct advantages and limitations. Data preprocessing and cleaning methods \cite{lee2018cleannet,lai2021tods,brodley1999identifying} attempt to identify and remove noisy samples before training, but often require substantial domain expertise and may inadvertently eliminate valuable edge cases or minority class examples. Noise-robust loss functions \cite{wang2019symmetric,zhang2018generalized} modify the training objective to reduce sensitivity to label errors\cite{ma2018dimensionality,han2018co}, yet these approaches typically rely on strong assumptions about noise distribution characteristics that may not hold in practice. Sample re-weighting\cite{jiang2018mentornet,patrini2017making} and curriculum learning techniques \cite{ren2018learning,liu2015classification} dynamically adjust the influence of training examples during optimization, but struggle with complex, asymmetric, and instance-dependent noise patterns commonly encountered in real-world scenarios \cite{xia2019anchor}. Meta-learning\cite{li2019learning,shu2019meta} approaches attempt to learn noise-robust representations, but require additional computational overhead and may not scale effectively to large-scale problems.

Machine unlearning emerges as a particularly promising paradigm for addressing data quality issues, originally developed to meet privacy compliance requirements by efficiently removing the influence of specific training samples without complete model retraining \cite{cao2015towards,bourtoule2021machine,graves2021amnesiac}. Unlike traditional data cleaning approaches that operate at the dataset level, unlearning techniques provide fine-grained control over individual sample influences within trained models. Recent theoretical and empirical investigations\cite{thudi2022necessity} have demonstrated that strategic application of unlearning methods can significantly enhance model robustness against noisy labels by selectively eliminating the impact of detrimental examples while preserving valuable generalizable knowledge \cite{golatkar2020eternal}. This selective forgetting capability offers computational efficiency advantages over complete retraining while maintaining model performance on high-quality data. Despite these advances, existing noise mitigation strategies face several critical limitations that constrain their practical applicability. Many approaches require explicit assumptions about noise characteristics, distributions, or generation processes that are difficult to verify in real-world settings. Others demand extensive manual intervention, domain-specific expertise, or substantial computational resources that may not be available in resource-constrained environments. 

To address these limitations, we propose a novel robust learning framework that synergistically combines attribution-based data\cite{sundararajan2017axiomatic, simonyan2013deep} quality assessment, discriminative neuron pruning\cite{han2015learning, louizos2017learning} , and targeted fine-tuning\cite{howard2018universal,ruder2019neural} to comprehensively address the challenges posed by both noisy input features and corrupted labels. Our approach leverages gradient-based attribution methods to quantify sample quality without requiring explicit noise distribution assumptions, employs neuron-level analysis to identify and remove noise-sensitive model components, and applies selective fine-tuning to restore and enhance model performance on high-quality data.
The key contributions of this work include:

\begin{enumerate}
\item \textbf{Attribution-guided Data Partitioning}: We utilize gradient-based attribution scores to reliably distinguish high-quality from noisy samples. Leveraging Gaussian mixture models allows for probabilistic clustering without imposing restrictive assumptions about the noise distributions.
\item \textbf{Discriminative Neuron Pruning}: We introduce a novel methodology for quantifying neuron sensitivity to noise as a linear regression problem based on neuron activations. This strategy enables precise identification and removal of neurons primarily influenced by noisy samples.
\item \textbf{Targeted Fine-Tuning}: After pruning, the network is fine-tuned exclusively on high-quality data subsets, effectively recovering and enhancing its generalization capability without incurring substantial computational costs.

\end{enumerate}

Our proposed framework addresses critical gaps in existing noise mitigation strategies by providing a unified approach that operates without explicit noise modeling requirements, scales effectively to large-scale models and datasets, and integrates seamlessly into standard deep learning workflows. The empirical results demonstrate significant improvements in model robustness and generalization across diverse noise scenarios, contributing valuable insights and practical tools for developing reliable deep learning systems in noisy real-world environments.

The remainder of this paper is systematically structured as: Section \ref{sec:algorithm} establishes the theoretical foundation by presenting essential preliminaries from noise-robust learning and machine unlearning. Building upon this framework, Section \ref{sec:main} elaborates our proposed methodology with rigorous algorithmic formulations. Section \ref{sec:experiments} provides empirical validation through controlled experiments, including comparative analysis with baselines and ablation studies. Section \ref{sec:con} summarizes our main findings and their implications for robust learning under noise. It also highlights key limitations and future research paths such as adaptive pruning and deployment in high-noise domains.

\section{Preliminaries}
\label{sec:algorithm}

In this section, we introduce the formal notation and core concepts that serve as the foundation for our proposed methodology.  We first formulate the supervised learning problem in the presence of noisy labels, then review gradient-based attribution techniques for measuring sample quality, describe neuron pruning strategies that target spurious representations, and finally summarize the fine-tuning paradigm used to adapt pruned networks.

\subsection{Supervised Learning} 

Let $\mathcal{X}$ denote the input space and $\mathcal{Y}$ the output space. In supervised learning, we aim to learn a function $f: \mathcal{X} \rightarrow \mathcal{Y}$ that maps inputs to outputs. Given a dataset $\mathcal{D} = \{(x_i, y_i)\}_{i=1}^n$ where $x_i \in \mathcal{X}$ and $y_i \in \mathcal{Y}$, the standard approach is to find parameters $\theta$ of a model $f_\theta$ that minimize the empirical risk:
\begin{equation}
\min_{\theta} \frac{1}{n} \sum_{i=1}^n \mathcal{L}(f_\theta(x_i), y_i)
\end{equation}

where $\mathcal{L}: \mathcal{Y} \times \mathcal{Y} \rightarrow \mathbb{R}^+$ is a loss function measuring the discrepancy between predicted and actual outputs.

In practical scenarios, the observed dataset is often contaminated by measurement noise and data corruption that can affect both input features and target outputs. Specifically, we can represent the observed dataset as $\mathcal{D}_{\text{obs}} = {(\tilde{x}_i, \tilde{y}_i)}_{i=1}^n$, where $\tilde{x}_i = x_i + \delta_i$ and $\tilde{y}_i = y_i + \epsilon_i$. Here, $\delta_i$ and $\epsilon_i$ represent input and output noise respectively, which may follow unknown distributions, exhibit adversarial characteristics, or arise from systematic measurement errors. This dual contamination presents significant challenges as the corruption can manifest in various forms: additive noise, multiplicative distortions, missing values, or adversarially crafted perturbations.

Standard empirical risk minimization applied to such corrupted data often leads to suboptimal parameter estimation and poor generalization performance. The degradation is particularly pronounced when the noise magnitude is significant, affects a substantial portion of the dataset, or when the noise characteristics differ between training and deployment environments. Input noise can cause the model to learn spurious correlations and reduce robustness to distributional shifts, while output noise leads to inconsistent supervision signals that impede convergence to the true underlying function.

\subsection{Attribution Methods}

Attribution methods provide a principled framework for quantifying the contribution of individual input features to a model's predictions, serving as fundamental tools for model interpretability and explainability. In the context of learning with noisy observations, we leverage attribution techniques to identify and characterize samples that may contain corrupted inputs or outputs, exploiting the hypothesis that such samples exhibit distinctive attribution patterns.

Formally, for input space $\mathcal{X} \subseteq \mathbb{R}^d$, an attribution method $\phi: \mathcal{X} \times \mathcal{Y} \times \mathcal{F} \rightarrow \mathbb{R}^d$ maps an input-output pair $(x, y)$ and a model $f \in \mathcal{F}$ to a $d$-dimensional vector in the input space, where each element quantifies the contribution of the corresponding feature to the model's prediction. This mapping enables us to decompose the model's decision-making process and identify the most influential features, applicable across different tasks and model types.

The landscape of attribution methods encompasses diverse approaches, broadly categorized into gradient-based and perturbation-based techniques \cite{Zeiler2014,Lundberg2017}. Gradient-based methods leverage the model's differentiability to compute feature importance through backpropagation, while perturbation-based approaches systematically modify inputs to observe changes in model behavior. Among these methods, we focus on Integrated Gradients (IG) \cite{Sundararajan2017}, which satisfies several desirable axiomatic properties including sensitivity, implementation invariance, and completeness, making it particularly suitable for robust analysis.

For a model $f$, input $x$, and baseline $x'$ (typically chosen as a zero vector or neutral reference point), Integrated Gradients is mathematically defined as:
\begin{equation}
\phi_{\text{IG}}(x, y, f) = (x - x') \odot \int_{\alpha=0}^{1} \nabla_x f(x' + \alpha(x - x')) \, d\alpha
\end{equation}
where $\nabla_x f(x)$ denotes the gradient of $f$ with respect to $x$, and $\odot$ represents element-wise multiplication. The integral captures the accumulated gradients along the straight-line path from the baseline to the input, providing a principled attribution that satisfies the completeness axiom.

In practice, the integral is approximated using Riemann sum numerical integration with $m$ steps:
\begin{equation}
\phi_{\text{IG}}(x, y, f) \approx (x - x') \odot \frac{1}{m} \sum_{k=1}^m \nabla_x f\left(x' + \frac{k}{m}(x - x')\right)
\end{equation}

The resulting attribution scores provide valuable insights into sample quality and data integrity. Our key insight is that samples containing noise---whether in inputs, outputs, or both---tend to exhibit anomalous attribution patterns compared to high-quality examples. Specifically, noisy samples often require the model to rely on spurious or irrelevant features to accommodate the corruption, leading to attribution distributions that deviate from the expected patterns of high-quality data. This deviation manifests as unusual magnitude distributions, unexpected feature importance rankings, or inconsistent attribution patterns across similar samples.

\subsection{Neural Network Pruning}

Neural network pruning is a model compression technique that systematically removes redundant connections or neurons from trained networks to reduce computational complexity while preserving performance on the target task. While traditional pruning methods primarily focus on efficiency gains through parameter reduction, recent advances have demonstrated that pruning can serve broader purposes, including the selective removal of specific learned representations and the enhancement of model robustness.

Consider a feedforward neural network (FNN) $f_{\theta}$ with $L$ layers and parameter set $\theta$. Let $W^l \in \mathbb{R}^{d_{l-1} \times d_l}$ denote the weight matrix connecting layer $l-1$ to layer $l$, where $d_l$ denotes the number of neurons in layer $l$. The forward propagation through layer $l$ for input $x$ is given by:
\begin{equation}
h^l = \sigma\left((W^l)^T h^{l-1} + b^l\right), \quad l = 1, \dots, L
\label{eq:layer_output}
\end{equation}
where $\sigma$ denotes a non-linear activation function, $b^l \in \mathbb{R}^{d_l}$ is the bias vector, and $h^0 = x$ represents the network input.

Structured pruning, particularly neuron-level pruning, involves the targeted elimination of individual computational units by zeroing their associated parameters. Specifically, pruning neuron \( j \) in layer \( l \) entails setting all of its incoming connections from layer \( l{-}1 \) to zero:
\begin{equation}
W^l_{j,:} = \mathbf{0}, \quad b^l_j = 0, \quad l = 1, \dots, L 
\end{equation}
This operation ensures that neuron \( j \) in layer \( l \) receives no input from the preceding layer, effectively disabling its activation in the forward pass.

The selection of neurons for pruning has been approached through various criteria in the literature. Magnitude-based methods \cite{han2015learning} identify less important parameters by examining weight magnitudes, operating under the assumption that smaller weights contribute less to network function. Sensitivity-based approaches \cite{molchanov2019importance} evaluate neuron importance by measuring the impact of their removal on network performance, typically through gradient analysis or direct performance assessment. Information-theoretic methods \cite{theis2018faster} leverage concepts such as Fisher information to quantify parameter importance based on the curvature of the loss landscape around the current parameter configuration.

Beyond computational efficiency, pruning has emerged as a powerful tool for understanding and modifying learned representations. The selective removal of neurons can reveal the internal organization of neural networks and provide insights into how different components contribute to various aspects of the learning task. This capability is particularly relevant in scenarios where networks may learn spurious correlations or overfit to noise, as targeted pruning can potentially mitigate these issues by removing the responsible computational pathways.
\subsection{Fine-tuning}

Fine-tuning is a transfer learning paradigm that adapts a pre-trained neural network to a new task or dataset by continuing the training process with carefully controlled parameter updates. This approach leverages the learned representations from the original training while allowing the model to specialize for the target domain, typically employing reduced learning rates to prevent catastrophic forgetting of useful pre-trained features.

Formally, consider a pre-trained model $f_{\theta}$ with initial parameter configuration $\theta_0$, and let $\mathcal{D}' = \{(x'_i, y'_i)\}_{i=1}^{n'}$ denote the target dataset for adaptation. The fine-tuning process optimizes the parameters $\theta$ by minimizing a regularized empirical risk objective:
\begin{equation}
\min_{\theta} \frac{1}{|\mathcal{D}'|} \sum_{(x'_i, y'_i) \in \mathcal{D}'} \mathcal{L}(f_{\theta}(x'_i), y'_i) + \lambda \Omega(\theta, \theta_0)
\end{equation}
where $\mathcal{L}$ represents the task-specific loss function (e.g., cross-entropy for classification or mean squared error for regression), $\Omega(\theta, \theta_0)$ is a regularization term that penalizes excessive deviation from the pre-trained weights, and $\lambda > 0$ controls the trade-off between task adaptation and parameter conservation.

The regularization component $\Omega(\theta, \theta_0)$ plays a crucial role in maintaining the stability of fine-tuning. A widely adopted choice is $\ell_2$-regularization:
\begin{equation}
\Omega(\theta, \theta_0) = \|\theta - \theta_0\|_2^2
\end{equation}
which encourages the fine-tuned parameters to remain in the vicinity of their pre-trained values, thereby preserving beneficial representations while allowing controlled adaptation to the new task.

The fine-tuning process typically incorporates several practical considerations to ensure effective transfer learning. These include the use of reduced learning rates compared to training from scratch, layer-wise learning rate scheduling that applies different rates to different network depths, and early stopping mechanisms that monitor validation performance to prevent overfitting to the target dataset. The choice of fine-tuning strategy—whether updating all parameters, only the final layers, or specific subsets—should be primarily task-driven, balancing computational efficiency and model performance based on the target task's requirements.

\section{Method}\label{sec:main}
\subsection{Problem setup}   
 
We investigate supervised learning with a neural network model $f_\theta: \mathcal{X} \rightarrow \mathcal{Y}$ parameterized by $\theta$, where $\mathcal{X}$ and $\mathcal{Y}$ denote the input and output spaces, respectively. The model is trained on a dataset $\mathcal{D} = \{(\hat{\mathbf{x}}_m, \hat{\mathbf{y}}_m)\}_{m=1}^M$, where each $(\hat{\mathbf{x}}_m,\hat{\mathbf{y}}_m)$ represents an input-output pair. In practical scenarios, such datasets frequently contain noise-corrupted samples that degrade model performance. While conventional approaches primarily enhance data robustness through preprocessing techniques to build systems resilient to imperfect data under complex quality conditions, our approach distinctively leverages the intrinsic dynamic properties of neural networks during training to identify high-fidelity samples.

We formally decompose dataset $\mathcal{D}$ into two disjoint subsets: $\mathcal{D}_r$ (high-quality data) and $\mathcal{D}_n$ (noise-corrupted data). The presence of $\mathcal{D}_n$ causes the learned model $f_\theta$ to deviate from the optimal function that would be obtained using exclusively $\mathcal{D}_r$. Our method employs $f_\theta$ to systematically partition $\mathcal{D}$ into $\mathcal{D}_r$ and $\mathcal{D}_n$, subsequently evaluates $f_\theta$ using these identified subsets to selectively prune noise-sensitive neurons, and finally fine-tunes $f_\theta$ on $\mathcal{D}_r$ to enhance generalization performance.

\begin{figure}[htbp]
    \centering
    \includegraphics[width=0.8\textwidth]{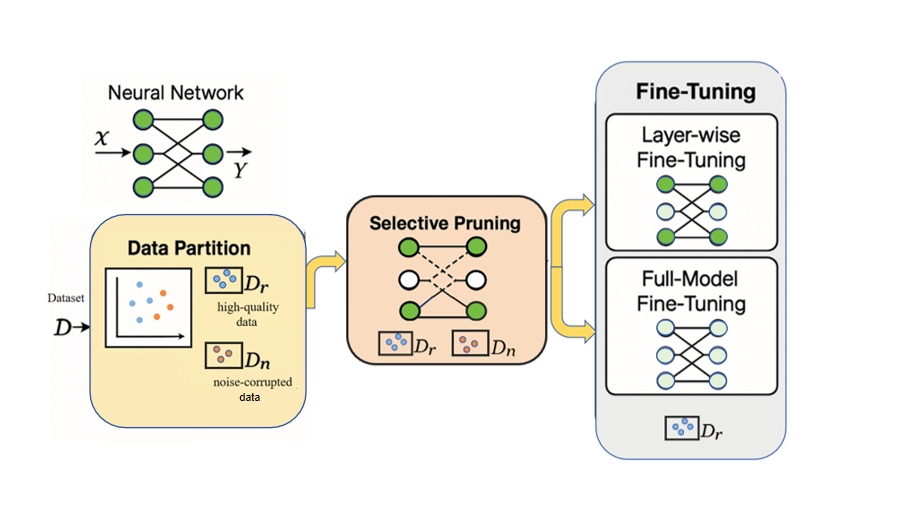}
    \caption{The overall algorithm flow. In the data, blue symbols represent high-quality data, while orange symbols indicate noise-corrupted data. In the neural network, green symbols denote trained neurons, white symbols represent neurons without information, and light leaf green symbols indicate neurons that will be trained. }
    \label{fig:overall}
\end{figure}

Our research objectives are threefold: 
(1) to develop an efficient algorithm for partitioning the dataset into $\mathcal{D}_r$ and $\mathcal{D}_n$ through analysis of $f_\theta$'s learning dynamics; 
(2) to establish a principled methodology for identifying and selectively pruning neurons that disproportionately contribute to learning from noisy data, utilizing both $\mathcal{D}_r$ and $\mathcal{D}_n$; and
(3) to optimize a fine-tuning procedure for the pruned model architecture on the high-quality subset $\mathcal{D}_r$ to restore and potentially enhance model performance and generalization capabilities. The overall algorithm flow is shown in Figure ~\ref{fig:overall}.

\subsection{Data partition}\label{sec:data}
To partition the dataset $\mathcal{D}$ into ${\mathcal{D}}_r$ and ${\mathcal{D}}_n$, we leverage neural attribution scores to quantify the contribution of individual training samples to the model's decision-making process. Within the model $f_\theta$, we employ a FNN architecture to define an attribution function that assigns a comprehensive quality score to each sample.
\subsubsection{Attribution Computation}
To quantify each sample's influence on model decisions, we compute attribution scores through gradient-based sensitivity analysis. For the model \( f_\theta \) with \( L \) layers, let \( \mathbf{f}^{l}_m \in \mathbb{R}^N \) denote the activation vector at layer \( l \) for the \( m \)-th sample, where \( N \) represents the number of neurons in that layer.
Let \(\mathbf{y}_m = f_{\theta}(\hat{\mathbf{x}}_m)\in \mathbb{R}^D\), and \( y_m^{d} \) denotes the $d$-th dimension of the output. We calculate the gradient of the model's output with respect to the activation \( \mathbf{f}^{l}_m \), denoted as \( \nabla_{\mathbf{f}^{l}_m} y_m^d \).
The attribution score for the \( m \)-th sample at layer \( l \) for output dimension \( d \) is formally defined as:
\begin{equation}\label{eq:attribution}
\mathbf{A}^{l}_{m,d} = \left| \mathbf{f}^{l}_{m} \odot \nabla_{\mathbf{f}^{l}_{m}} y_m^d \right|,
\end{equation}
To quantify the contribution of individual neurons in deep neural networks, we define $\mathbf{A}^{l,n}_{m,d}$ as the attribution score of the $m$-th neuron in layer $l$ for the $d$-th output dimension given the $n$-th input sample. To obtain a precise per-neuron importance measure for each sample, we denote $\mathbf{A}^{l}_{m,d} = \{\mathbf{A}^{l,n}_{m,d}\}_{n=1}^N$ and compress the attribution tensor through max-pooling across output dimensions.
This operation is mathematically defined as:
\begin{equation}
\hat{\mathbf{{A}}}^{l,n}_{m} = \max_{1 \leq d \leq D} \mathbf{A}^{l,n}_{m,d}
\end{equation}
Each element $\hat{\mathbf{{A}}}^{l,n}_{m} $ represents the most significant influence of neuron $n$ in layer $l$ on any output dimension for sample $m$.
\subsubsection{Attribution-Based Clustering via  Gaussian Mixture Model}
After computing the sample-wise attribution scores, we partition the dataset using probabilistic clustering techniques. Among various clustering methodologies evaluated, the Gaussian Mixture Model (GMM)     \cite{pearson1894contributions} demonstrated superior performance due to its capacity to capture non-spherical cluster shapes through covariance modeling, provide soft assignment probabilities, and accommodate varying cluster densities.
The GMM models the attribution vectors $\mathbf{a}_m = \{\hat{\mathbf{A}}^{l,1:N}_{m}\} \in \mathbb{R}^{N}$, where $m=1,\dots,M$, as a weighted superposition of $K$ Gaussian components:
\begin{equation}\label{eq:gmm_mixture}
p(\mathbf{a}_m) = \sum_{k=1}^{K} \pi_k \mathcal{N}(\mathbf{a}_m \mid \boldsymbol{\mu}_k, \boldsymbol{\Sigma}_k)
\end{equation}
where $\pi_k$ denotes the mixing coefficient (subject to $\sum_{k=1}^K \pi_k = 1$), $\boldsymbol{\mu}_k \in \mathbb{R}^N$ and $\boldsymbol{\Sigma}_k \in \mathbb{R}^{N \times N}$ represent the mean vector and covariance matrix of component (k), respectively, characterizing the central tendency and inter-neuron correlation structure of attribution scores.
For this binary partitioning scenario ($K=2$), We standardize all attribution vectors to follow a standard normal distribution $\mathcal{N}(0,1)$, initialize parameters via k-means++ \cite{arthur2006k} algorithm, and optimize the model parameters using Expectation-Maximization (EM) \cite{dempster1977maximum}.
The posterior cluster assignment probability for sample $m$ is given by:
\begin{equation}\label{eq:gmm_assignment}
\gamma_k(\mathbf{a}_m) = \frac{\pi_k \mathcal{N}(\mathbf{a}_m \mid \boldsymbol{\mu}_k, \boldsymbol{\Sigma}_k)}{\sum_{j=1}^K \pi_j \mathcal{N}(\mathbf{a}_m \mid \boldsymbol{\mu}_j, \boldsymbol{\Sigma}_j)}
\end{equation}
The final partitions are determined by classifying samples into high-quality data
\[\mathcal{D}_r = {\mathbf{a}_m \mid \arg\max_k \gamma_k(\mathbf{a}_m) = 1}\]
and noise-corrupted data 
\[\mathcal{D}_n = {\mathbf{a}_m \mid \arg\max_k \gamma_k(\mathbf{a}_m) = 2}.\]

\subsection{Selective Pruning}  
\label{sec:selective_prune}
To systematically identify influential neurons for pruning, we implement regression-based sensitivity analysis utilizing features extracted from a pre-trained model \( f_\theta \) and the partitioned dataset \( \mathcal{D} = \{(\hat{\mathbf{x}}_m, \hat{\mathbf{y}}_m)\}_{m=1}^M \) with corresponding binary quality labels \( z_m \in \{0,1\} \), where \( z_m = 1 \) denotes high-quality samples in ${\mathcal{D}}_r$ and \( z_m = 0 \) denotes noise-corrupted samples in ${\mathcal{D}}_n$.

Let \( \mathbf{f}_m^{(l)} = [f_{m,1}^{(l)}, f_{m,2}^{(l)}, \dots, f_{m,N}^{(l)}]^\top \in \mathbb{R}^N \) denote the activation vector at layer \( l \) for sample \( m \), where \( N \) represents the number of neurons in layer \( l \). We formulate and solve a least-squares linear regression problem to predict the sample's quality label \( z_m \) from these activations:
\begin{equation}
\min_{\mathbf{T}^{l}, u^{l}} \sum_{m=1}^{M} \left( z_m - {\mathbf{T}^{l}}^\top \mathbf{f}_m^{l} - u^{l} \right)^2,
\end{equation}
where \( \mathbf{T}^{l} = [T_1^{l}, T_2^{l}, \dots, T_N^{l}]^\top \in \mathbb{R}^N \) represents the vector of learned regression coefficients, and \( u^{l} \in \mathbb{R} \) is the intercept term.

For each neuron \( n \) in layer \( l \), we define a comprehensive sensitivity score $s_n$ that quantifies its influence on distinguishing between high-quality and noise-corrupted samples:
\begin{equation}
  s_n = |T_n^{l}| + \lambda|u_n^{l}|
\end{equation}
where $\lambda$ is a hyperparameter that modulates the relative importance of the weights versus the bias term in the sensitivity calculation.

To execute the pruning operation, we identify the top-$\alpha$ ($\alpha \in (0,1)$) proportion of neurons $N_{\text{prune}} = \{n_j \mid s_j \geq s_{(\lceil(1-\alpha)N\rceil)}\}$ ranked in descending order by their sensitivity scores. The pruning procedure systematically zeros out both the incoming weights and biases of the identified neurons in layer $l$:
\begin{equation}
\forall n_j \in N_{\text{prune}}: \mathbf{W}^{l}[j,:] \gets \mathbf{0},\quad \mathbf{b}^{l}[j] \gets 0
\end{equation}
where $\mathbf{W}^{l} \in \mathbb{R}^{N_{l} \times N_{l-1}}$ and $\mathbf{b}^{l} \in \mathbb{R}^N$ represent the weight matrix and bias vector of layer $l$, respectively. This selective pruning process yields a refined model $f_{\theta'}$ with enhanced robustness to noise-corrupted training samples.

\subsection{Fine-Tuning}

Following selective pruning, the model undergoes a fine-tuning phase exclusively on the high-quality subset ${\mathcal{D}}_r$. This step is crucial for restoring the performance deterioration caused by pruning and reinforcing the learning of clean patterns.

We obtained new parameters $\theta$ by fine-tuning the model.
\begin{equation} 
\min_{\theta} \frac{1}{|{\mathcal{D}}_r|} \sum_{(\bar{x_i}, \bar{y_i}) \in {\mathcal{D}}_r} \mathcal{L}(f_{\theta}(\bar{x_i}), \bar{y_i}) + \lambda_{\text{reg}} \|\theta - \theta’\|_2^2
\end{equation}

where $\mathcal{L}$ is the task-specific loss function (e.g., cross-entropy for classification), $\theta'$ is the parameters after prune after ~\ref{sec:selective_prune} , and $\lambda_{\text{reg}}$ is a regularization coefficient that controls the deviation from the pruned model. This regularization term prevents catastrophic forgetting of useful information learned before pruning.

During the fine-tuning stage, we designed two strategies to compare their impact on final performance:

\paragraph{Layer-wise Fine-Tuning}
Only parameters $\theta^{l}$ of layer $l$ are updated; others paramters ($\theta^{\text{frozen}} = \theta' - \theta^{l}$) remain fixed. The objective becomes:

\begin{equation} \label{eq:layer_ft}
\boldsymbol{\theta}^{l*} = \arg\min_{\boldsymbol{\theta}^{l}} \left( \frac{1}{|\mathcal{D}_r|} \sum_{(\bar{x_i}, \bar{y_i}) \in \mathcal{D}_r} \mathcal{L}\big(f_{\theta^{l}, \theta^{\text{frozen}}}(\bar{x_i}), \bar{y_i}\big) + \lambda_{\text{reg}} \|\theta^{l} - \theta_p^{l}\|_2^2 \right)
\end{equation}

Here, $\theta_p^{l}$ are the pruned parameters of layer $l$. The output is $f^{\text{layer}}_{\theta}$.

\paragraph{Full-Model Fine-Tuning}
All parameters $\boldsymbol{\theta^*}$ are updated jointly:

\begin{equation} \label{eq:full_ft}
\boldsymbol{\theta}^* = \arg\min_{\boldsymbol{\theta^*}} \left( \frac{1}{|\mathcal{D}_r|} \sum_{(\bar{x_i}, \bar{y_i}) \in \mathcal{D}_r} \mathcal{L}\big(f_{\theta^{*}}(\bar{x_i}), \bar{y_i}\big) + \lambda_{\text{reg}} \|\theta^{*} - \theta'\|_2^2 \right)
\end{equation}

The output is $f^{\text{full}}_{\theta}$.

By comparing the performance of $f_{\theta}$, $f^{\text{layer}}_{\theta}$, and $f^{\text{full}}_{\theta}$ on the test set, we can effectively demonstrate the validity of our method.
The complete methodology is presented in Algorithm~\ref{alg:rla} 
\begin{algorithm}[H]
\caption{Robust Learning via Attribution-Based Pruning (RLAP)}
\label{alg:rla}
\begin{algorithmic}[1]
\Require 
    Training dataset $\mathcal{D} = \{(\hat{\mathbf{x}}_m, \hat{\mathbf{y}}_m)\}_{m=1}^M$ \\
    Pretrained FNN model $f_\theta$ with $L$ layers \\
    Target layer index $l \in \{1,...,L\}$ \\
    Pruning ratio $\alpha$ \\
    Regularization coefficient $\lambda_{\text{reg}}$
    
\Ensure Refined model $f_{\theta}^{\text{final}}$

\State \textbf{Phase 1: Data Partitioning}
\State Pretrained FNN model $f_\theta$ with $L$ layers
\For{each sample $(\hat{\mathbf{x}}_m, \hat{\mathbf{y}}_m) \in \mathcal{D}$}
    \State Forward pass to compute layer-$l$ activations: $\mathbf{f}_m^l = f_\theta^{l}(\hat{\mathbf{x}}_m) \in \mathbb{R}^N$ and the output \(\mathbf{y}_m = f_{\theta}(\hat{\mathbf{x}}_m)\in \mathbb{R}^D\)
    \State Backpropagate to compute gradients of $d- $th dimension of output: $\nabla_{\mathbf{f}_m^l} y_m^d = \frac{\partial y_m^d}{\partial \mathbf{f}_m^l}$
    \State Compute attribution matrix: 
    $\mathbf{A}^{l}_{m,d} = \left| \mathbf{f}^{l}_{m} \odot \nabla_{\mathbf{f}^{l}_{m}} y_m^d \right|$  
    \Comment{$\mathbf{f}^{l}_{m}$: feature vector of sample $m$ at layer $l$;  
             $\nabla_{\mathbf{f}^{l}_{m}} y_m^d$: gradient of output $y_m^d$ w.r.t. features;  
            $\mathbf{A}^{l}_{m,d}$: attribution scores} 
    \State Max-pool across outputs: $\hat{\mathbf{A}}_m^{l,n} = \{\max_{1\leq d \leq D} A_{m,d}^{l,n}\}_{n=1}^N$
\EndFor

\State Fit GMM with $K=2$ components on $\{\mathbf{a}_m\}_{m=1}^M$ with $\mathbf{a}_m=\hat{\mathbf{A}}_m^{l,1:N}$
\State Partition $\mathcal{D}$ into $\mathcal{D}_r$ and $\mathcal{D}_n$ using Equation~\eqref{eq:gmm_assignment}

\State \textbf{Phase 2: Layer-$l$ Selective Pruning}
\State Extract layer-$l$ activations $\{\mathbf{f}_m^l\}_{m=1}^M$ and quality labels $\{z_m\}_{m=1}^M$
\State Solve linear regression: $\min_{\mathbf{T}^l,u^l} \sum_m (z_m - {\mathbf{T}^l}^\top\mathbf{f}_m^l - u^l)^2$
\For{each neuron $n$ in layer $l$}
    \State Compute sensitivity score $s_n = |T_n^l| + \lambda|u_n^l|$
\EndFor
\State Select top-$\alpha$ neurons $N_{\text{prune}}$ by $s_n$
\State Zero weights and biases for layer $l$:
\State $\mathbf{W}^l[j,:] \gets \mathbf{0}, \mathbf{b}^l[j] \gets 0$ for $j \in N_{\text{prune}}$
\State Obtain pruned model $f_{\theta'}$

\State \textbf{Phase 3: Fine-Tuning}
\State Option A: Layer-$l$ fine-tuning using Equation~\eqref{eq:layer_ft}
\State Option B: Full-model fine-tuning using Equation~\eqref{eq:full_ft}
\State Return the fine-tuning model $f_{\theta}^{\text{layer}}$ or  $f_{\theta}^{\text{full}}$.
\end{algorithmic}
\end{algorithm}


The comprehensive experimental validation of our method will be presented in Section~\ref{sec:experiments}, where we expect to observe: (i) consistent performance improvement over baseline models, (ii) greater robustness to varying noise levels compared to retraining approaches, and (iii) computational efficiency gains from our selective pruning strategy.

\section{Experiment}
\label{sec:experiments}
To validate our proposed Robust Learning via Attribution-Based Pruning (RLAP) framework, we conducted comprehensive evaluations across two distinct application domains: computer vision (CIFAR-10 image classification) and speech recognition (Speech Commands keyword spotting). Our test results only consider noise in features for classification tasks, but our method can easily be extended to tasks where both features and labels have noise.
In the above scenario, we opted for the classical method and network design to train the prediction, and for the corresponding dataset. To simulate realistic low-quality training data, we artificially introduced noise to half of the training samples. Our experiments employed a hybrid CNN-FNN architecture; however, optimization via attribution-based pruning was selectively applied to the FNN layer due to the contextual complexity of CNN layers, making FNN a suitable initial testbed. We acknowledge that extending our framework to encompass CNN layers represents a promising avenue for future research.


Subsequently, we designated the trained model for each task as the initial model baseline. To demonstrate the effectiveness of our proposed approach, our method includes these two strategies for comparison: layer-wise fine-tuning ($f^{\text{layer}}_{\theta}$, L-FT) and full-model fine-tuning($f^{\text{full}}_{\theta}$, F-FT) across all experimental tasks. Furthermore, to validate the efficacy of our proposed data partition methodology, we incorporated an additional baseline involving a retrained model. The retrained model differs from the initial model solely in the composition of the training set, where the full dataset is replaced with the refined subset $D_{r}$ obtained through the data differentiation process described in Section~\ref{sec:data}. This comparison enables us to isolate and assess the contribution of our data curation approach independently from the pruning strategy outlined in Section~\ref{sec:selective_prune}.

Following the methodology established in Section~\ref{sec:main}, all experiments were implemented using Algorithm~\ref{alg:rla} in PyTorch with a fixed pruning ratio of $\alpha=0.15$. This pruning ratio was determined through preliminary ablation studies, which verified that $\alpha=0.15$ effectively removes the majority of data-sensitive neurons from the initial model while maintaining optimal performance.

\subsection{Computer vision}
\subsubsection{Dataset and Experimental Setup}
\label{sec:computervision}
We evaluate our method on CIFAR-10, a widely adopted benchmark in computer vision research. This dataset comprises 60,000 color images ($32\times32$ pixels) equally distributed across 10 object classes, with a standard split of 50,000 training and 10,000 test samples. CIFAR-10 serves as an ideal testbed due to its manageable computational scale, balanced class distribution, and established role in evaluating CNN architectures, hyperparameter optimization, and novel training methodologies. Beyond classification tasks, this dataset facilitates comprehensive studies on transfer learning, semi-supervised learning, data augmentation, and robustness against noisy inputs—making it particularly relevant for our noise-robust learning framework.


We adopt resnet-18 as the backbone architecture for all experiments. We conduct comprehensive performance comparisons across four key evaluation metrics:

$$
\begin{aligned}
\text{Accuracy} &= \frac{\text{TP} + \text{TN}}{\text{TP} + \text{TN} + \text{FP} + \text{FN}}, \\
\text{Precision} &= \frac{\text{TP}}{\text{TP} + \text{FP}}, \\
\text{Recall} &= \frac{\text{TP}}{\text{TP} + \text{FN}} &\quad\\
\text{F1-score} &= 2 \times \frac{\text{Precision} \times \text{Recall}}{\text{Precision} + \text{Recall}},
\end{aligned}
$$

where TP (True Positives) is the number of correctly predicted positive samples, TN (True Negatives) is the number of correctly predicted negative samples, FP (False Positives) is the number of negative samples incorrectly predicted as positive and FN (False Negatives) is the number of positive samples incorrectly predicted as negative. And in our experiment, Recall is equal with Accuracy so we have not show it in table.





We evaluate these metrics at varying training scales and measure the per-epoch computational time. We compare our proposed methods with the results of the company's experiment. ( see Table  ~\ref{tab:finetune_results_eng})

To establish competitive baselines, we compare against prior work by Pochinkov et al.~\cite{pochinkov2024dissecting}, which proposed four neuron-scoring methods for machine unlearning. The baseline importance functions are defined as:

\begin{align*}
I_{\text{abs}}(\mathcal{D}, n) &= \frac{1}{|\mathcal{D}|} \sum_{m=1}^{M} |\mathbf{f}_m^{l}(n)| \\
I_{\text{rms}}(\mathcal{D}, n) &= \sqrt{\frac{1}{|\mathcal{D}|} \sum_{m=1}^{M} (\mathbf{f}_m^{l}(n))^2} \\
I_{\text{freq}}(\mathcal{D}, n) &= \frac{1}{|\mathcal{D}|} \cdot \left|\{m \in \{1,...,M\} \mid \mathbf{f}_m^{l}(n) > 0\}\right| \\
I_{\text{std}}(\mathcal{D}, n) &= \sqrt{\frac{1}{|\mathcal{D}|} \sum_{m=1}^{M} \left(\mathbf{f}_m^{l}(n) - \bar{\mathbf{f}}^{l}(n)\right)^2}
\end{align*}

where $n$ is the neuron in layer $l$ of model $f_\theta$, $\mathbf{f}_m^{l}(n)$ denotes the activation of neuron $n$ in layer $l$ for input sample $\hat{\mathbf{x}}_m$ and $\bar{\mathbf{f}}^{l}(n) = \frac{1}{M}\sum_{m=1}^{M} \mathbf{f}_m^{l}(n)$ is the mean activation of neuron $n$.

The scoring function for comparing neuron behavior between high-quality  ($\mathcal{D}_{\text{r}}$) and noisy ($\mathcal{D}_{\text{n}}$) subsets is:

\begin{equation}
\text{Score}(n, \mathcal{D}_{\text{r}}, \mathcal{D}_{\text{n}}) = \frac{I(\mathcal{D}_{\text{n}}, n)}{I(\mathcal{D}_{\text{r}}, n) + \epsilon}
\end{equation}

where $\epsilon > 0$ is a small constant for numerical stability

Our novel pruning method (Section~\ref{sec:selective_prune}) outperforms these baselines in post-pruning accuracy (Table ~\ref{tab:pruning_comparison}). We further validate robustness through noise-level experiments, demonstrating consistent performance across varying noise conditions (Table ~\ref{tab:noise_results_eng}).

All implementations use PyTorch with NVIDIA A10 GPUs. Training epochs: 60 epochs for Initial Model and Retrain Model, 30 epochs for fine-tuning variants (L-FT/F-FT), the results are evaluated independently on the test set.

\subsubsection{Results and Analysis}
\begin{figure}[htbp]
\centering
\includegraphics[width=0.8\textwidth]{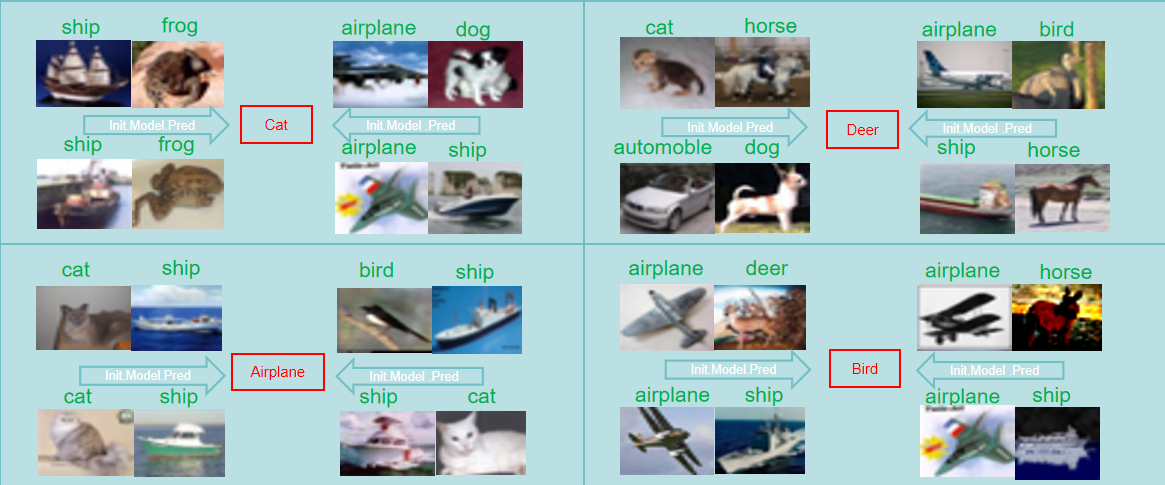}
\caption{Representative samples misclassified by the initial model but correctly predicted by both our fine-tuning strategies. Green labels indicate correct predictions by our method. Red labels indicate wrong predictions by initial model.}
\label{fig:example1}
\end{figure}
Figure~\ref{fig:example1} provides visual evidence of our method's capability to rectify classification errors made by the baseline model. The displayed samples—misclassified by the initial model but correctly predicted by both fine-tuning variants—demonstrate the effectiveness of our noise-robust learning approach in recovering discriminative features that were corrupted during standard training procedures.

\begin{table}[!htbp]
\centering
\small
\caption{Comparative Analysis of Fine-tuning Strategies Across Different Training Scales}
\label{tab:finetune_results_eng}
\begin{tabular}{@{} l *{4}{>{\centering\arraybackslash}p{1.5cm}} p{1.5cm} @{}}
\toprule
\multirow{2}{*}{\textbf{Train Size}} & \multirow{2}{*}{\textbf{Metric}} & \multicolumn{4}{c}{\textbf{Method}} \\
\cmidrule(lr){3-6}
 & & \rotatebox{0}{\parbox{1.5cm}{\centering Initial\\Model}} 
 & \rotatebox{0}{\parbox{1.5cm}{\centering L-FT}} 
 & \rotatebox{0}{\parbox{1.5cm}{\centering F-FT}} 
 & \rotatebox{0}{\parbox{1.5cm}{\centering Retrain\\Model}} \\
\midrule

\multirow{4}{*}{50k} 
 & Accuracy & 0.6944 & 0.7508 & 0.8020 & 0.7299 \\
 & Precision & 0.6962 & 0.7546 & 0.8075 & 0.7450 \\
 & F1-score & 0.6962 & 0.7500 & 0.8024 & 0.7298 \\
 & Time (s) & 20.48 & 7.98 & 14.44 & 15 \\
\cmidrule(lr){1-6}

\multirow{4}{*}{25k}
 & Accuracy & 0.6196 & 0.6818 & 0.7253 & 0.6616 \\
 & Precision & 0.6710 & 0.6835 & 0.7257 & 0.6879 \\
 & F1-score & 0.6129 & 0.6812 & 0.7228 & 0.6569 \\
 & Time (s) & 11.68 & 3.1 & 5.55 & 5.43 \\
\cmidrule(lr){1-6}

\multirow{4}{*}{12.5k}
 & Accuracy & 0.5619 & 0.6067 & 0.6124 & 0.5828 \\
 & Precision & 0.5857 & 0.6112 & 0.6226 & 0.6238 \\
 & F1-score & 0.5525 & 0.6070 & 0.6108 & 0.5884 \\
 & Time (s) & 6.87 & 2.02 & 3.63 & 4.77 \\
\bottomrule
\end{tabular}

\vspace{2mm}
\end{table}

\begin{table}[!htbp]
\centering
\scriptsize
\setlength{\tabcolsep}{5pt}
\renewcommand{\arraystretch}{1.0}
\caption{Comparative Analysis of Pruning Methods Across Training Scales}
\label{tab:pruning_comparison}
\begin{tabular}{@{} l @{\hspace{4pt}} c @{\hspace{4pt}} *{15}{>{\centering\arraybackslash}p{0.7cm}} @{}}
\toprule
\multirow{2}{*}{\textbf{Size}} & \multirow{2}{*}{\textbf{Metric}} & \multicolumn{3}{c}{\textbf{Abs}} & \multicolumn{3}{c}{\textbf{Rms}} & \multicolumn{3}{c}{\textbf{Freq}} & \multicolumn{3}{c}{\textbf{Std}} & \multicolumn{3}{c}{\textbf{Ours}} \\
\cmidrule(lr){3-5} \cmidrule(lr){6-8} \cmidrule(lr){9-11} \cmidrule(lr){12-14} \cmidrule(lr){15-17}
 & & Pruned & L-FT & F-FT & Pruned & L-FT & F-FT & Pruned & L-FT & F-FT & Pruned & L-FT & F-FT & Pruned & L-FT & F-FT \\
\midrule

\multirow{4}{*}{50k} 
 & Acc. & 0.4707 & 0.7353 & 0.7966 & 0.5073 & 0.7586 & 0.7894 & 0.5480 & 0.7742& 0.7853 & 0.4139 & 0.7526 & 0.7988 & 0.2423 & 0.7508 & 0.8020 \\
 & Prec. & 0.4874 & 0.7360 & 0.7976 & 0.6498 & 0.7579 & 0.7984 & 0.7284 & 0.7729 & 0.7867 & 0.5801 & 0.7547 & 0.7896 & 0.3610 & 0.7546 & 0.8075 \\
 & F1 & 0.4140 & 0.7443 & 0.7966 & 0.4449 & 0.7579 & 0.7894 & 0.5425 & 0.7731 & 0.7827 & 0.4139 & 0.7526 & 0.7965 & 0.1388 & 0.7500 & 0.8024 \\
\addlinespace[2pt]

\multirow{4}{*}{25k}
 & Acc. & 0.5492 & 0.2731 & 0.1285 & 0.5737 & 0.6759 & 0.6872 & 0.6169 & 0.4683 & 0.2738 & 0.5587 & 0.6755 & 0.7073 & 0.1771 & 0.6818 & 0.7253 \\
 & Prec. & 0.6647 & 0.3199 & 0.4678 & 0.6908 & 0.6742 & 0.6878 & 0.6808 & 0.5747 & 0.4358 & 0.6151& 0.6772 & 0.7216 & 0.0800 & 0.6835 & 0.7257 \\
 & F1 & 0.6294 & 0.2029 & 0.0757 & 0.5565 & 0.6741 & 0.6857 & 0.6203 & 0.4382 & 0.1942 & 0.5148 & 0.6753 & 0.7077 & 0.0938 & 0.6812 & 0.7228 \\
\addlinespace[2pt]

\multirow{4}{*}{12.5k}
 & Acc. & 0.5567 & 0.5859 & 0.6216 & 0.5104 & 0.5939 & 0.6040 & 0.2552 & 0.5515 & 0.6104 & 0.4735 & 0.5704 & 0.6296 & 0.1000 & 0.6067 & 0.6124 \\
 & Prec. & 0.5929 & 0.5911 & 0.6223 & 0.5468 & 0.5930 & 0.6093 & 0.1738 & 0.5574 & 0.5531 & 0.5442 & 0.5660& 0.6345 & 0.0100 & 0.6112 & 0.6266 \\
 & F1 & 0.5490 & 0.5859 & 0.6196 & 0.5133 & 0.5924 & 0.6035 & 0.1213 & 0.5531 & 0.6104 & 0.4416 & 0.5666 & 0.6309 & 0.0182 & 0.6070 & 0.6108 \\
\bottomrule
\end{tabular}

\vspace{2mm}
\end{table}

\begin{table}[!htbp]
\centering
\small
\caption{Comparative Analysis of Fine-tuning Strategies Under Different Noise Levels (Train Size=50k)}
\label{tab:noise_results_eng}
\begin{tabular}{@{} l >{\centering\arraybackslash}p{1.5cm} >{\centering\arraybackslash}p{1.5cm} >{\centering\arraybackslash}p{1.5cm} >{\centering\arraybackslash}p{1.5cm} >{\centering\arraybackslash}p{1.5cm} @{}}
\toprule
\multirow{2}{*}{\textbf{Noise Level}} & \multirow{2}{*}{\textbf{Metric}} & \multicolumn{4}{c}{\textbf{Method}} \\
\cmidrule(lr){3-6}
 & & \rotatebox{0}{\parbox{1.5cm}{\centering Initial\\Model}} 
 & \rotatebox{0}{\parbox{1.5cm}{\centering L-\\ FT}} 
 & \rotatebox{0}{\parbox{1.5cm}{\centering F-\\ FT}} 
 & \rotatebox{0}{\parbox{1.5cm}{\centering Retrain\\Model}} \\
\midrule

\multirow{4}{*}{level=2}
 & Accuracy & 0.7128 & 0.7741 & 0.7934 & 0.7127 \\
 & Precision & 0.7309 & 0.7749 & 0.7969 & 0.7499 \\
 & F1-score & 0.7126 & 0.7735 & 0.7945 & 0.7123 \\
 & Time (s) & 20.48 & 12.28 & 21.86 & 22.72 \\
\cmidrule(lr){1-6}

\multirow{4}{*}{level=3}
 & Accuracy & 0.7220 & 0.7754 & 0.7966 & 0.6855 \\
 & Precision & 0.7442 & 0.7749 & 0.7968 & 0.7257 \\
 & F1-score & 0.7237 & 0.7746 & 0.7951 & 0.6806 \\
 & Time (s) & 20.48 & 12.16 & 21.53 & 22.84 \\
\cmidrule(lr){1-6}

\multirow{4}{*}{level=4}
 & Accuracy & 0.7079 & 0.7761 & 0.7904 & 0.7058 \\
 & Precision & 0.7269 & 0.7796 & 0.7932 & 0.7468 \\
 & F1-score & 0.7062 & 0.7758 & 0.7902 & 0.7086 \\
 & Time (s) & 20.48 & 10.35 & 18.55 & 18.12 \\
\cmidrule(lr){1-6}

\multirow{4}{*}{level=5}
 & Accuracy & 0.6818 & 0.7754 & 0.7954 & 0.7074 \\
 & Precision & 0.7370 & 0.7742 & 0.7969 & 0.7448 \\
 & F1-score & 0.6857 & 0.7743 & 0.7952 & 0.7131 \\
 & Time (s) & 20.45 & 6.94 & 12.69 & 12.45 \\
\cmidrule(lr){1-6}

\multirow{4}{*}{level=6}
 & Accuracy & 0.6992 & 0.7654 & 0.7923 & 0.7039 \\
 & Precision & 0.7033 & 0.7672 & 0.7934 & 0.7301 \\
 & F1-score & 0.6923 & 0.7656 & 0.7915 & 0.7053 \\
 & Time (s) & 20.48 & 8.71 & 15.59 & 18.0 \\
\cmidrule(lr){1-6}

\multirow{4}{*}{level=7}
 & Accuracy & 0.7292 & 0.7575 & 0.7925 & 0.6307 \\
 & Precision & 0.7360 & 0.7585 & 0.7946 & 0.6770 \\
 & F1-score & 0.7295 & 0.7568 & 0.7922 & 0.6192 \\
 & Time (s) & 20.48 & 6.60 & 12.12 & 14.78 \\
\cmidrule(lr){1-6}

\multirow{4}{*}{level=8}
 & Accuracy & 0.6944 & 0.7508 & 0.8020 & 0.7299 \\
 & Precision & 0.6962 & 0.7546 & 0.8075 & 0.7450 \\
 & F1-score & 0.6962 & 0.7500 & 0.8024 & 0.7298 \\
 & Time (s) & 20.48 & 7.98 & 14.44 & 15 \\
\cmidrule(lr){1-6}

\multirow{4}{*}{level=9}
 & Accuracy & 0.7150 & 0.7657 & 0.8006 & 0.6986 \\
 & Precision & 0.7482 & 0.7648 & 0.8034 & 0.7562 \\
 & F1-score & 0.7183 & 0.7643 & 0.8012 & 0.6945 \\
 & Time (s) & 20.61 & 6.67 & 12.21 & 12.02 \\
\bottomrule
\end{tabular}

\vspace{2mm}
\footnotesize
Note: Training size fixed at 50k for all experiments.
\end{table}

\textbf{Superior Performance of Full-Model Fine-tuning}: As demonstrated in Table~\ref{tab:finetune_results_eng}, Full-model Fine-tuning achieves superior performance and enhanced capacity recovery across all data scales. At the complete 50k training size, Full-model Fine-tuning attains 80.20\% accuracy—representing a substantial 10.76\% absolute improvement over the noise-corrupted initial model (69.44\%) and a significant 9.29\% enhancement over conventional retraining (72.99\%). This dual improvement confirms that our pruning strategy effectively preserves critical discriminative features while eliminating noise-sensitive components.

\textbf{Computational Efficiency}: Our specialized Fine-tuning achieves rapid performance restoration with minimal computational overhead. As evidenced in Table~\ref{tab:finetune_results_eng}, Full-model Fine-tuning requires only 14.44s per epoch—28\% less time than retraining (15s per epoch)—while delivering superior accuracy (80.20\% vs. 72.99\%). Remarkably, this performance gain is achieved with half the training epochs (30 vs. 60), demonstrating our method's accelerated convergence enabled by precise neuron pruning.

\textbf{Data-efficient Generalization}: As training data availability decreases (Table~\ref{tab:finetune_results_eng}), our method maintains significant performance advantages. With only 12.5k samples (25\% of the full dataset), Full-model Fine-tuning achieves 61.24\% accuracy—a substantial 5.05\% absolute improvement over retraining (58.28\%). This result highlights our method's ability to effectively leverage limited high-quality data for robust model training.

\textbf{Attribution-Guided Pruning Effectiveness}: Table~\ref{tab:pruning_comparison} demonstrates our attribution-guided pruning strategy's unique capability to identify and eliminate noise-sensitive neurons, resulting in superior final model performance. The results reveal three key advantages: (1) Our method exhibits the most significant initial accuracy drop (24.23\% at 50k scale), indicating precise targeting of neurons most adversely affected by noisy samples. (2) Following Full-model Fine-tuning, our approach achieves 80.20\% accuracy—outperforming all alternatives, including the second-best Standard deviation method's Full-model Fine-tuning (79.88\%) by 1.32\%, despite requiring less training time. (3) The performance advantage becomes more pronounced at smaller training scales, with our method maintaining a 5.8\% accuracy lead (72.53\% vs. 68.72\%) over RMS's Full-model Fine-tuning at 25k scale, demonstrating robust scalability across different data regimes.

\textbf{Noise-Adaptive Performance}: Table~\ref{tab:noise_results_eng} demonstrates that our data partitioning algorithm achieves consistent advantages as noise levels increase, primarily through improved model accuracy due to enhanced separation of high-quality and noisy data. At noise level 2, Full-model Fine-tuning outperforms the Retrain Model by 8.07\% (79.34\% vs. 71.27\%). This performance gap widens to 10.20\% at level 9 (80.06\% vs. 69.86\%), confirming that higher noise levels enhance the discriminative power of neurons in distinguishing $\mathcal{D}_r$ (high-quality data) from $\mathcal{D}_n$ (noisy data). L-FT demonstrates the highest computational efficiency, reducing training time by 40.0–67.8\% compared to retraining (e.g., 6.67s vs. 12.02s at level 9). Full-model Fine-tuning maintains high accuracy while being up to 47.2\% faster than retraining, achieving an optimal performance-efficiency balance.

The synergistic combination of precise attribution-based pruning followed by targeted Fine-tuning achieves dual objectives: (1) efficient removal of noise-induced interference and (2) enhanced recovery of discriminative capacity beyond standard approaches. This advantage is particularly evident under extreme conditions (high noise levels or limited data availability), where conventional methods deteriorate while our framework maintains robust performance.

\subsection{Speech Recognition}
\subsubsection{Dataset and Experimental Setup}
To validate our method's generalizability across domains, we evaluate on speech recognition using the Speech Commands dataset (10-class subset), a standard benchmark for keyword spotting tasks. This dataset contains 35,000 one-second audio clips sampled at 16kHz, covering 10 core vocal commands ("yes", "no", etc.) along with additional silence and unknown samples. The dataset's constrained vocabulary and uniform temporal duration make it particularly suitable for evaluating noise-robust learning approaches in speech applications.

Our experimental setup employs a CNN+FFN hybrid architecture. For this evaluation, we adopt a custom 80-20 training-testing partition rather than the standard partition to better demonstrate our approach's effectiveness across different data distributions.
Consistent with Section ~\ref{sec:computervision}, we assess performance using three metrics: Accuracy, Precision, F1-score and the per-epoch computational time. What's more towards this task, We adopt Top-3 Accuracy to evaluate model performance, which checks if the ground-truth label is within the top-3 predicted classes. Formally, for a dataset with $M$ samples, it is computed as:

\begin{equation}
\text{Top-3 Accuracy} = \frac{1}{M} \sum_{i=1}^{M} \mathbb{I}\left(y_i \in \{\text{Top-3 predictions for } x_i\}\right)
\label{eq:top3}
\end{equation}

\subsubsection{Results and Analysis}

\begin{figure}[htbp]
    \centering
    \includegraphics[width=0.8\textwidth]{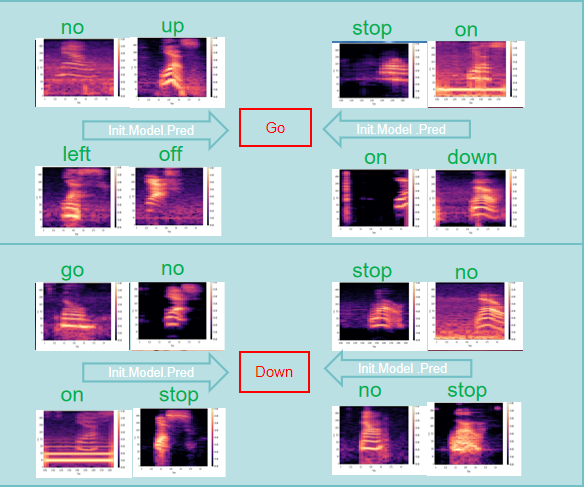}
    \caption{Representative samples misclassified by the initial model but correctly predicted by both our fine-tuning strategies. Green labels indicate correct predictions by our method. Red labels indicate wrong predictions by initial model.}
    \label{fig:example2}
\end{figure}
Figure~\ref{fig:example2} presents visualization results using Mel-spectrogram preprocessing (64 bins, 25ms window, 10ms hop length). We display representative samples from two categories that demonstrate the diversity of audio features. The visualization reveals that while the initial model incorrectly predicts certain audio samples, our proposed method achieves correct classification, indicating substantial room for improvement in the baseline approach.

\begin{table}[!htbp]
\centering
\caption{Comparative Analysis in Speech Commons}
\label{tab:commons_show1}
\begin{tabular}{@{}lcccccc@{}}
\toprule
\multirow{2}{*} & \multirow{2}{*}{\textbf{Stage}} & \multicolumn{4}{c}{\textbf{Performance Metrics}} & \multirow{2}{*}{\textbf{Time (s)}} \\
\cmidrule(lr){3-6}
 & & \textbf{Accuracy (\%)} & \textbf{Precision (\%)} & \textbf{F1 (\%)} & \textbf{Top-3 Acc (\%)} & \\
\midrule

\multirow{4}{*}
 & Initial Model & 64.20 & 64.03 & 64.02 &89.92 & 6.90 \\
 & L-FT & 66.94 & \textbf{76.34} & \textbf{77.01} & 90.45 & 1.32\\
 & F-FT&  \textbf{71.21} & 72.90 & 71.92&\textbf{93.41} & 2.85\\
 & Retrain Model &  65.37& 65.31 & 65.27& 90.38 & 3.23 \\
\addlinespace

\bottomrule
\end{tabular}

\vspace{2mm}
\footnotesize
Note: All metrics are reported in percentage (\%). Time represents average training time per epoch. Best results are highlighted in bold. Time recorded per epoch on NVIDIA Tesla T4 GPU.
\end{table}

Table~\ref{tab:commons_show1} demonstrates consistent performance advantages across all evaluation metrics. Accuracy improvements: Full-model Fine-tuning achieves 71.21\% accuracy—representing a 6.84\% improvement over the initial model (64.20\%) and a 5.84\% enhancement over retraining (65.37\%). The substantial 7.01\% F1-score improvement (71.92\% vs. 64.02\%) confirms enhanced noise immunity. Computational efficiency: L-FT delivers a 3.23× speedup (1.32s vs. 3.23s per epoch) compared to retraining while maintaining superior precision (76.34\% vs. 65.31\%). Full-model Fine-tuning achieves an optimal performance-speed balance at 2.85s per epoch. Prediction reliability: Our method's 93.41\% top-3 accuracy indicates consistent prediction reliability even when the correct label is not the highest-ranked prediction—a crucial characteristic for practical speech interface applications.

This comprehensive validation on speech recognition data strengthens our method's claim as a domain-agnostic solution for noise-robust learning across diverse application areas.

\section{Conclusion}\label{sec:con}

In this study, we have introduced a robust deep learning framework designed explicitly to address the pervasive issue of noisy training data through the integration of machine unlearning principles, attribution-guided data partitioning, discriminative neuron pruning, and targeted fine-tuning. Our approach initially employs gradient-based attribution methods combined with Gaussian Mixture Models to effectively distinguish high-quality samples from noise-corrupted ones without imposing restrictive assumptions on noise characteristics. Subsequently, we apply a regression-based neuron sensitivity analysis to selectively identify and prune neurons highly susceptible to noise interference. This strategic pruning is followed by selective fine-tuning—either layer-wise or full-model—focused exclusively on clean, high-quality data, incorporating regularization to mitigate potential knowledge degradation.
Empirical validation across diverse domains, including CIFAR-10 image classification and Speech Commands keyword spotting, underscores the effectiveness and generalizability of our framework. 

Future research directions include investigating adaptive pruning mechanisms, online unlearning methodologies to enhance computational efficiency, and expanding applicability to semi-supervised and unsupervised learning scenarios. Additionally, further exploration into theoretical convergence and generalization guarantees under varying noise conditions, as well as validation on larger-scale benchmarks and domains prone to high noise levels—such as medical imaging and natural language processing—will provide deeper insights and broader applicability. Ultimately, our work contributes a practical and theoretically grounded approach for achieving reliable and robust deep neural networks in noisy real-world environments.

 \bibliographystyle{unsrt}
 \bibliography{references}
\end{document}